\documentclass[conference]{IEEEtran}
\IEEEoverridecommandlockouts
\usepackage{cite}
\usepackage{amsmath,amssymb,amsfonts}
\usepackage{graphicx}
\usepackage{textcomp}
\usepackage{xcolor}

\usepackage{graphicx}
\usepackage{lscape}
\usepackage{multirow}
\usepackage{tabularx}
\usepackage{booktabs}       
\usepackage{xcolor}
\usepackage{placeins}
\usepackage{hyperref}

\usepackage{algorithm,algorithmicx}
\usepackage[noend]{algpseudocode}

\hypersetup{breaklinks=true}

\makeatletter
\let\MYcaption\@makecaption
\makeatother

\usepackage[font=footnotesize]{subcaption}

\makeatletter
\let\@makecaption\MYcaption
\makeatother

\def\BibTeX{{\rm B\kern-.05em{\sc i\kern-.025em b}\kern-.08em
    T\kern-.1667em\lower.7ex\hbox{E}\kern-.125emX}}

\iffalse

\newcommand{\pk}[1]{\textcolor{red}{\small [pk: #1]}}
\newcommand{\piotrm}[1]{\textcolor{blue}{\small [pm: #1]}}
\newcommand{\kucil}[1]{\textcolor{magenta}{\small [łk: #1]}}
\newcommand{\kc}[1]{\textcolor{green}{\small [kc: #1]}}

\else

\newcommand{\pk}[1]{}
\newcommand{\piotrm}[1]{}
\newcommand{\kucil}[1]{}
\newcommand{\kc}[1]{}

\fi

\renewcommand{\S}{\mathcal{S}}
\newcommand{\A}{\mathcal{A}}
\renewcommand{\P}{\mathcal{P}}
\newcommand{\R}{\mathcal{R}}
\renewcommand{\H}{\mathcal{H}}
\newcommand{\Havg}{\H_\mathrm{avg}}
\newcommand{\E}{\mathbb{E}}
\newcommand{\Esp}{\E_{s' \sim \P(\cdot | s, a)}}
\newcommand{\argmax}{\mathrm{argmax}}

\def\withrefs{0}

\newcommand{\robustref}[2]{\if\withrefs{\autoref{#1}}\else{#2}}
\newcommand{\appendixref}[1]{\if\withrefs{Appendix~\ref{#1}}\else{the Appendix}}
    
\begin{document}

\title{Planning and Learning using Adaptive Entropy Tree Search}

\iffalse

\author{\IEEEauthorblockN{Anonymous Authors}}

\else

\author{\IEEEauthorblockN{Piotr Kozakowski}
\IEEEauthorblockA{
\textit{Faculty of Mathematics, Informatics} \\
\textit{and Mechanics} \\
\textit{University of Warsaw} \\
Warsaw, Poland \\
p.kozakowski@mimuw.edu.pl}
\and
\IEEEauthorblockN{Mikołaj Pacek}
\IEEEauthorblockA{
\textit{Faculty of Mathematics, Informatics} \\
\textit{and Mechanics} \\
\textit{University of Warsaw} \\
Warsaw, Poland \\
m.pacek2@student.uw.edu.pl}
\and
\IEEEauthorblockN{Piotr Miłoś}
\IEEEauthorblockA{
\textit{Institute of Mathematics} \\
\textit{of the Polish Academy of Sciences} \\
Warsaw, Poland \\
pmilos@impan.pl}
}

\fi

\maketitle

\begin{abstract}

\pk{abstract proposal}

Recent breakthroughs in Artificial Intelligence have shown that the combination of tree-based planning with deep learning can lead to superior performance. We present Adaptive Entropy Tree Search (ANTS) - a novel algorithm combining planning and learning in the maximum entropy paradigm. Through a comprehensive suite of experiments on the Atari benchmark we show that ANTS significantly outperforms PUCT, the planning component of the state-of-the-art AlphaZero system. ANTS builds upon recent work on maximum entropy planning methods - which however, as we show, fail in combination with learning. ANTS resolves this issue to reach state-of-the-art performance. We further find that ANTS exhibits superior robustness to different hyperparameter choices, compared to the previous algorithms. We believe that the high performance and robustness of ANTS can bring tree search planning one step closer to wide practical adoption.

\end{abstract}


\kucil{General remarks:
\begin{itemize}
    \item Probably a matter of taste (I think we differ slightly with pm as it comes to this), but I would rather avoid philosophical statements (e.g. "the main goals of AI..."): a) it brings little to the table b) noboody cares what we think / we are not hinton, lecun, bengio c) it takes space and makes it longer for the reader to get to the point of the paper. \pk{I wanted to highlight why stuff we talk about is important, esp for the reviewers that don't see it a priori; that's why I wanted to start with this paragraph}
    \item I think the main point of the paper should be: "ANTS is the first maximum entropy MCTS method showing competitive performance". ATM it is slighlty hidden. The intro should be very clear about this (in the text and in the contributions). The experiment section you have to very clearly highlight this point (first exp. section should stand out and be very clear about this; something like "this section presents the main result of this paper...").
    \item This paper seems slightly out of focus. It has multiple ambitions: a) speculate about the goals, state and important directions of AI b) be a survey: at length explain RL, MCTS with its intricacies, maxent RL with intricacies, and on top of that provide lengthly related work. 
    (TBH I can relate, but I think this approach is hurtful to the paper, here is why:) This is slightly confusing for me as the reader: a) I am not focused b) it is only at page 2.5 that I start learning about ANTS c) as I arrive to ANTS I am already tired and slightly overloaded with different terms, equations, etc., and the next item. \pk{but then how to make the paper self-contained? I've seen reviewers complaining about that}
    \item ANTS description is watered down and I can't really say what it does in the end (could also be that I read it later in the evening). What can be done with this: a) shorten the into b) shorten Background and make it lighter c) Put all the elements of ANTS in one place d) Put the pseudocode in the paper (now should be more space available) \pk{agreed, will add the pseudocode if space allows}
    \item I would avoid adverbs, such as "very..." (it pops up several times in the paper, and each time it sounds not very professional).
    \item I would avoid '-' in the text (or use it sparingly); use "," or ";" instead; \pk{why?}
    \item When referring to appendix, provide a number. Similarly with sections. \pk{appendix will sadly be removed}
    \item I didn't use grammarly, so you know.
\end{itemize}
More concrete remarks follow.
}

\section{Introduction} \label{sec:intro}


Planning and learning is a fundamental component of intelligent behavior in both biological and artificial agents \cite{hassabis2017neuroscience, russell2002artificial, DBLP:journals/eor/BengioLP21}. 
\kucil{more positive (bez "hands"); focus on AI, e.g. From the latter point of view}
On the one hand, planning can discover high-quality decisions but falls short in large state spaces. On the other hand, learning can harness
powerful pattern matching techniques and discover useful state
representations, hence significantly reducing the search space \cite{alphazero,Agostinelli:2019aa}.

Recently, Monte Carlo tree search (MCTS) planning methods \cite{mcts-survey, uct}, which incorporate learning, played a major role in some of the breakthroughs in deep reinforcement learning (RL) and in the real world domains, ranging from playing complex strategic games \cite{mogo}, through testing security systems \cite{mcts-security}, physics simulations \cite{mcts-physics}, production management \cite{mcts-production}, to creative content generation \cite{mcts-creative}.
Consequently, there exists a large body of research aiming to 
improve the performance of MCTS \cite{alphazero,muzero,mcts-as-rpo,effzero}. Importantly, MCTS algorithms are modular, allowing to incorporate a broad spectrum of ideas.

One such prominent idea is the Principle of Maximum Entropy (PME) \cite{maxent-phd,maxent-collective}, 
which has improved the performance and increased the robustness of RL methods \cite{maxent-is-the-answer, sql, sac}, and was 
combined with MCTS in \cite{ments, tents}.
In particular, the latter papers introduce MENTS and TENTS algorithms, respectively, demonstrating how to utilize pre-trained value networks in maximum entropy planning, and also providing theoretical guarantees.
However, the successful scaling of MENTS and TENTS to the online learning setting has been an open problem. 
In this paper, we show how to close this gap, by introducing Adaptive Entropy Tree Search (ANTS), the first maximum entropy MCTS method showing competitive performance in online learning on the Atari benchmark \cite{ale}.

\kucil{Here describe (maybe in 1-2 sentences) what \emph{is} ANTS}

ANTS combines three algorithmic enhancements: (a) temperature adaptation, (b) leaf evaluation using raw Q-values, and (c) pseudoreward shaping. The eponymous temperature adaptation introduces a useful reparameterization of the algorithm: instead of controlling the temperature of the action selection distribution, we control the average entropy of that distribution across the nodes in the tree. This parameterization, as we show experimentally, is significantly more robust to different hyperparameter choices. It also has a more intuitive interpretation: the Shannon entropy of a discrete distribution can be regarded as a logarithm of its \textit{weighted average branching factor} \cite{branching-factor}. As such, by setting the level of entropy, we control the breadth of the search tree.

\section{Background}

Reinforcement learning (RL) problems are typically formalized by Markov decision processes (MDPs). An MDP is a tuple $\langle \mathcal{S, A, P, R}, \gamma \rangle$, where $\S, \A$ are the state and action spaces, $\P(\cdot| s, a)$ is the transition distribution, $\R$ is a scalar reward function and $\gamma \in [0, 1)$ is the discount factor. The agent's behavior is described by the \textit{policy} $\pi(\cdot | s), s \in \S$. 

Maximum entropy RL introduces the objective 
\begin{equation*} 
    J(\pi) = \E_\pi \left[ \sum_{t = 0}^\infty \gamma^t \left[ \R(s_t, a_t) + \tau \H(\pi(\cdot | s_t)) \right] \right]
\end{equation*}
where $\tau > 0$ is a temperature parameter and $\H$ is Shannon entropy:
\begin{equation}
    \label{eq:shannon}
    \H(\pi(\cdot | s_t)) = -\sum_{a \in \mathcal{A}} \pi(a | s_t) \log \pi(a | s_t).
\end{equation}
\cite{sql} derives a backup operator, which can be used to solve entropy-regularized RL problems\footnote{In the limit as $\tau \searrow 0$ we recover the classical Bellman backup.}:
%
\begin{align} \label{eq:sqi}
\begin{split}
    Q(s, a) & \leftarrow \R(s, a) + \gamma \Esp V(s') \\
    V(s) & \leftarrow \tau \log \sum_{a \in \A} \exp \left( \frac{1}{\tau} Q(s, a) \right).
\end{split}
\end{align}
The fixed point of this backup results in $Q^\star, V^\star$. The policy $\pi^\star(\cdot | s) \propto \exp\left(\frac{1}{\tau}  Q^\star(s, a)  \right)$ is optimal with respect to $J$.

%
\paragraph{Monte Carlo Tree Search (MCTS)}
MCTS is a class of planning algorithms in MDPs with discrete action spaces. Below we describe its version related in this paper, referring to \cite{mcts-survey} for a general discussion. MCTS iteratively builds a planning tree where nodes correspond to states and edges to (state, action) pairs. It consists of four phases: \textbf{Selection}, in which a path from the root to a leaf is selected,
\textbf{Expansion}, in which all successors of this leaf are added to the tree as new leaves,
\textbf{Simulation}, in which the values of the new leaves are estimated, 
and \textbf{Backpropagation}, in which the value estimates of the nodes on the path are updated, starting from the bottom.



%
AlphaZero \cite{alphazero} augments vanilla MCTS with neural network heuristics trained in a self-improvement loop. AlphaZero uses the PUCT selection strategy, in which at state $s$ the next edge/action is chosen to maximize
%
%
%
\begin{equation} \label{eq:puct}
    \mathrm{PUCT}(s, a) = Q(s, a) + c~\pi_\textrm{prior}(a | s) \frac{\sqrt{N(s)}}{N(s, a) + 1},
\end{equation}
where $c>0$ is a hyperparamer, $\pi_\textrm{prior}(a | s)$ is a prior policy, $N(s)$ is the number of visits in state $s$ and $N(s, a)$ is the number of times action $a$ has been executed in state $s$. In the simulation phase, the value of a leaf is estimated using a value network. This network is trained to estimate the empirical state values. $\pi_\textrm{prior}$ is also a network, trained to predict $\frac{N(s, a)}{N(s)}$.


In this work we consider an expansion phase that adds all successors of a given node to the tree. These new leaves are initialized using a \textit{Q-network}. The $Q$-network is trained to estimate the empirical state-action values.

After planning is finished, an action is selected according to a stochastic policy $\pi$ (this approach follows \cite{muzero}):
\begin{equation}
    \label{eq:puct_sel}
     \pi(a)\propto N(s_{\textrm{root}}, a)^{{1}/{\tau_\textrm{sel}}},
    \end{equation}
where $\tau_\textrm{sel} \in (0, 1)$ is a temperature hyperparameter. The stochastic $\pi$ ensures diverse data for training. If exploration is not needed, i.e. when not training neural networks, we use $\argmax_{a}  N(\text{root}, a)$ (equivalent to $\tau_\textrm{sel} \searrow 0$).

\paragraph{MaxEnt MCTS} The maximum entropy backup \eqref{eq:sqi} can be adapted to MCTS, resulting in the MENTS algorithm \cite{ments}. In the selection phase the consecutive nodes are chosen according to the E3W sampling strategy:
\begin{align} \label{eq:e3w}
    \begin{split}
        \pi_\tau^\mathrm{e3w}(a | s) &= (1 - \lambda_s)~\pi_\tau(a | s) + \lambda_s \frac{1}{|\A|}, \\
        \pi_\tau(a | s) &\propto \exp \left( \frac{1}{\tau} Q(s, a) \right),
    \end{split}
\end{align}
where $\lambda_s = \frac{\epsilon |\A|}{\log (N(s) + 1)}, \epsilon >0$. In the simulation phase, leaves are initialized using:
\begin{align} \label{eq:tents_eval}
\begin{split}
Q(s, a) &\leftarrow \left( \hat{Q}(s, a) - \hat{V}(s) \right) / \tau_\textrm{init}, \\
\hat{V}(s) &= \tau \log \sum_{a \in \A} \exp \left( \frac{1}{\tau} \hat{Q}(s, a) \right),
\end{split}
\end{align}
where $\hat{Q}$ is a $Q$-network. In the backpropagation phase, node $Q$-values are updated according to \eqref{eq:sqi}. The algorithm described here utilizes Shannon entropy. It can be extended to other types of entropy; in particular \cite{tents} derive TENTS, a version using Tsallis entropy:
\begin{equation}
    \label{eq:tsallis}
    \H(\pi(\cdot | s_t)) = \frac{1}{2} \left( 1 - \sum_{a \in \mathcal{A}} \pi(a | s_t)^2\right ).
\end{equation}

Because optimizing the maximum-entropy RL objective with Tsallis entropy can result in sparse policies, this formulation is particularly beneficial in environments with large action spaces \cite{sparse-mdp}.


%
%
%
\section{Method} \label{sec:method}

\iffalse


We now detail our Adaptive Entropy Tree Search (ANTS), see  Algorithm~\ref{algo:ants}. It is coupled with online learning, see Algorithm~\ref{algo:learning}. Our learning loop is similar to the one used in \cite{alphazero} and shown. We emphasise that our algorithmic developments, presented below, are critical for making online learning feasible.

We use hyperparameters $\Havg, \tau_0, \tau_{\textrm{min}}, \tau_{\textrm{sel}}, \epsilon, n, k$ (described below).\piotrm{put somewhere information about possibility of using fixed $Q$. I think it is enough in experiments.}



\begin{algorithm}[ht]
\caption{Adaptive Entropy Tree Search.}
\label{algo:ants}

\begin{algorithmic}
\Function{ants}{$s_{\textrm{root}}, \Tilde{\tau}$}
\For{$i$ \textbf{in} $\{1,\ldots, n\}$}
\If{$i \mod k = 0$}
\State{$\tau \gets \max(\textsc{find\_root}(\mathcal{H}^+), \tau_\mathrm{min})$}
\State{$\Tilde{\tau} \gets \exp(\alpha \log \Tilde{\tau} + (1 - \alpha) \log \tau)$}
\State $\textsc{recalculate\_q}(s_{\textrm{root}}, \Tilde{\tau})$
\EndIf
\State $s_{\textrm{leaf}} \gets \textsc{selection}(s_{\textrm{root}}, \Tilde{\tau})$
\State children[$s_{\textrm{leaf}}$] $\gets \textsc{expansion}(s_{\textrm{leaf}})$
\State $~Q(s_{\textrm{leaf}}, a) \gets \hat{Q}(s_{\textrm{leaf}}, a; \tau), a \in \mathcal{A}$
\State $\textsc{backpropagation}(s_{\textrm{leaf}}, rewards, \Tilde{\tau})$
\EndFor
\State \textbf{sample} $a \sim \pi_{\Tilde{\tau} \cdot \tau_{\mathrm{sel}}}^\mathrm{e3w}(\cdot | s_{\textrm{root}})$
\State \textbf{return} $a, \Tilde{\tau}, Q(s_{\textrm{root}}, a; \Tilde{\tau})$
\EndFunction
\end{algorithmic}

\begin{algorithmic}
\Function{recalculate\_q}{$s$, $\tau$}
\For{$a \in \A$}
\State $\textsc{recalculate\_q}(\textrm{children}[$s$][$a$], \tau)$
\State $Q(s, a) \gets \textsc{backup}(s, a, \tau)$
\EndFor
\EndFunction
\end{algorithmic}

\end{algorithm}
ANTS adheres to the general template of (MaxEnt) MCTS, with $n$ planning steps for each environment move.  Importantly, $\tau$ tracks temperature, adapted every $k$ steps. Its update necessitates recalculation of $Q$ estimates in the MCTS tree. In the selection phase, we use the E3W policy \eqref{eq:e3w} sharpened by a separate temperature parameter $\tau_{\textrm{sel}}$. \texttt{BACKUP} uses the max entropy backup \eqref{eq:shaping}. \eqref{eq:shaping} is based on Shannon entropy, we evaluate our method also in a setting with Tsallis entropy. \piotrm{write about ANTS-T, ANTS-S in the experimental section. Othewise nobody will remember}



\begin{algorithm}[ht]
\caption{Selection and backpropagation phases.}
\label{algo:select}

\begin{algorithmic}
\Function{selection}{$s$, $\tau$}
\While{\textbf{not} $\textsc{is\_leaf}(s)$}
\State \textbf{sample} $a \sim \pi_\tau^\mathrm{e3w}(\cdot | s)$
\State $s \gets \textrm{children}[s][a]$
\EndWhile
\State \textbf{return} $s$
\EndFunction
\end{algorithmic}

\begin{algorithmic}
\Function{backpropagation}{$s$, $\tau$}
\While{\textbf{not} $\textsc{is\_root}(s)$}
\State $N(s) \gets N(s) + 1$
\State $Q(s, a) \gets \textsc{backup}(s, a, \tau)$
\State $s \gets \textrm{parent}[s]$
\EndWhile
\EndFunction
\end{algorithmic}

\end{algorithm}



\begin{algorithm}[ht]
\caption{Online learning.}
\label{algo:learning}

\begin{algorithmic}
\Function{learning\_loop}{}
\State $\textrm{replay\_buffer} \gets ()$
\While{\textbf{not} convergence}
\State episode $\gets \textsc{collect\_episode}$
\State $\textrm{replay\_buffer} \gets \textrm{replay\_buffer} \oplus (\textrm{episode})$
\State $\textrm{batch} \gets \textsc{replay\_sample}(\textrm{replay\_buffer})$
\State $\hat{Q} \gets \textsc{sgd}(L, \textrm{batch}, \hat{Q})$
\EndWhile
\EndFunction
\end{algorithmic}

\begin{algorithmic}
\Function{collect\_episode}{}
\State $s \gets s_0; \tau \gets \tau_0$; episode $\gets (s_0)$
\While{\textbf{not} $\textsc{is\_terminal}(s)$}
\State $a, \tau, q \gets \textsc{ants}(s, \tau)$
\State \textbf{sample} $s \sim \P(\cdot | s, a)$
\State episode $\gets$ episode $\oplus~(a, \tau, q, s)$
\EndWhile
\State \textbf{return} episode
\EndFunction
\end{algorithmic}

\end{algorithm}




\paragraph{Temperature adaptation} \label{sec:temp_adaptation}
In our method the temperature $\tau$ is adapted continuously. It may change even during a single episode to reflect exploration strategies. Formally, $\tau$ is varied so that the mean node entropy (over the nodes the MCTS tree $\S_\mathrm{tree}$) matches the prescribed value $\Havg$. This is achieved by finding a root of the following function:
\begin{align*}
    \H^+(\tau) &= \left[ \frac{1}{|\S_\mathrm{tree}|} \sum_{s \in \S_\mathrm{tree}} \H(\pi_\tau(\cdot | s)) \right] - \Havg, \\
    \textrm{where}~&\pi_\tau(a | s) \propto \exp \left( \frac{1}{\tau} Q(s, a) \right).
\end{align*}
In practice we use 
\begin{align*}
    & \tau \gets \max(\textsc{find\_root}(\mathcal{H}^+), \tau_\mathrm{min}) \\
    & \Tilde{\tau} \gets \exp(\alpha \log \Tilde{\tau} + (1 - \alpha) \log \tau), \quad \alpha \in (0,1)
\end{align*}
where for the sake of simplicity, \texttt{FIND\_ROOT} is implemented using the Brent algorithm \cite{brent}, $\tau_\mathrm{min}>0$ prevents numerical issues for $\tau \approx 0$. The moving average in the second line is intended to stabilize changes. 




\paragraph{Pseudoreward shaping} 
The $\textsc{backup}$ used in Algorithms~\ref{algo:ants} and~\ref{algo:select} is a modification of \eqref{eq:sqi}:
\begin{align} \label{eq:shaping}
    \begin{split}
        Q(s, a; \tau) & \leftarrow \R(s, a) + \gamma \Esp V(s'; \tau) \\
        V(s; \tau) & \leftarrow \tau \log \sum_{a \in \A} \exp \left( \frac{1}{\tau} Q(s, a; \tau) \right) - \tau\H_\mathrm{max}.
    \end{split}
\end{align}
where $\H_\mathrm{max}$ is the maximum possible entropy (for the given action space). The additional pseudoreward $- \tau\H_\mathrm{max}$ counterbalances the positive contributions due to entropy in \eqref{eq:sqi}. These contributions promote an inefficient depth-first tree exploration (as newly expanded leaf are assigned erroneously optimistic values), which we observed in experiments. We note that, as the subtracted term does not depend on the action, this shaping does not change the optimal policy obtained at infinite computational budget.




\paragraph{$Q$-value estimation} 
In the simulation phase, we initialize the expanded leaves using a network $\hat{Q}(s, a;\tau)$ which approximates soft $Q$-values at temperature $\tau$. Besides its simplicity, we found empirically this method much more effective than the original MaxEnt MCTS formulation \eqref{eq:tents_eval}.




%
\paragraph{Exploration} \label{sec:learning} 


During data collection in Algorithm \ref{algo:learning}, we use an exploration scheme similar to \eqref{eq:puct_sel}. Actions performed at each step in the environment are sampled from:
\begin{equation}
\label{eq:ants_sel}
\pi(a | s) = \pi_{\tau \cdot \tau_\textrm{sel}}^\mathrm{e3w}(a | s),
\end{equation}
where $\tau_\textrm{sel} \in (0, 1)$ is an action selection temperature parameter. It interpolates between the E3W \eqref{eq:e3w} and the argmax policy: $a(s) = \argmax_{a \in \A}~Q(s, a)$.

\else

We now describe our method, Adaptive Entropy Tree Search (ANTS). We follow the general template of MaxEnt MCTS, as shown in Algorithm~\ref{algo:ants}. The hyperparameters of the algorithm are $\Havg, \tau_0, \tau_{\textrm{min}}, \tau_{\textrm{sel}}, \alpha, \epsilon, n, k$.

\begin{algorithm}[ht]
\caption{Adaptive Entropy Tree Search.}
\label{algo:ants}

\begin{algorithmic}
\Function{ants}{$s_{\textrm{root}}, \Tilde{\tau}$}
\For{$i$ \textbf{in} $\{1\ldots n\}$}
\If{$i \mod k = 0$}
\State{$\tau \gets \max(\textsc{find\_root}(\mathcal{H}^+), \tau_\mathrm{min})$}
\State{$\Tilde{\tau} \gets \exp(\alpha \log \Tilde{\tau} + (1 - \alpha) \log \tau)$}
\State $\textsc{recalculate\_q}(s_{\textrm{root}}, \Tilde{\tau})$
\EndIf
\State $s_{\textrm{leaf}} \gets \textsc{selection}(s_{\textrm{root}}, \Tilde{\tau})$
\State children[$s_{\textrm{leaf}}$] $\gets \textsc{expansion}(s_{\textrm{leaf}})$
\State $\forall_{a \in \A}~Q(s_{\textrm{leaf}}, a) \gets \hat{Q}(s_{\textrm{leaf}}, a; \tau)$
\State $\textsc{backpropagation}(s_{\textrm{leaf}}, \Tilde{\tau})$
\EndFor
\State \textbf{sample} $a \sim \pi_{\Tilde{\tau} \cdot \tau_{\mathrm{sel}}}^\mathrm{e3w}(\cdot | s_{\textrm{root}})$
\State \textbf{return} $a, \Tilde{\tau}, Q(s_{\textrm{root}}, a; \Tilde{\tau})$
\EndFunction
\end{algorithmic}

\begin{algorithmic}
\Function{recalculate\_q}{$s$, $\tau$}
\For{$a \in \A$}
\State $\textsc{recalculate\_q}(\textrm{children}[$s$][$a$], \tau)$
\State $Q(s, a) \gets \textsc{backup}(s, a, \tau)$
\EndFor
\EndFunction
\end{algorithmic}

\end{algorithm}

At every timestep we make $n$ MCTS iterations. Every $k$ iterations we adapt the temperature and recalculate the $Q$-values in the tree. Temperature adaptation is described in Section~\ref{sec:temp_adaptation}. $Q$-value recalculation involves a custom backup described in Section~\ref{sec:shaping}. The selection and backpropagation phases are shown in Algorithm~\ref{algo:select} below. The expansion phase is omitted for clarity. The simulation phase is inlined as a call to the $Q$-network; this is described in detail in Section~\ref{sec:value_estimation}. Finally, we select an action by sampling from the E3W policy \eqref{eq:e3w} sharpened by a separate temperature parameter $\tau_{\textrm{sel}}$, described in Section~\ref{sec:learning}.

\begin{algorithm}[ht]
\caption{Selection and backpropagation phases.}
\label{algo:select}

\begin{algorithmic}
\Function{selection}{$s$, $\tau$}
\While{\textbf{not} $\textsc{is\_leaf}(s)$}
\State \textbf{sample} $a \sim \pi_\tau^\mathrm{e3w}(\cdot | s)$
\State $s \gets \textrm{children}[s][a]$
\EndWhile
\State \textbf{return} $s$
\EndFunction
\end{algorithmic}

\begin{algorithmic}
\Function{backpropagation}{$s$, $\tau$}
\While{\textbf{not} $\textsc{is\_root}(s)$}
\State $N(s) \gets N(s) + 1$
\State $Q(s, a) \gets \textsc{backup}(s, a, \tau)$
\State $s \gets \textrm{parent}[s]$
\EndWhile
\EndFunction
\end{algorithmic}

\end{algorithm}

In the selection phase, we choose the consecutive nodes until we reach a leaf by sampling from the E3W policy \eqref{eq:e3w}. In the backpropagation phase we recalculate the $Q$-values of the nodes on the path from the leaf to the root using the backup described in Section~\ref{sec:shaping}.

In addition to the evaluation using pretrained Q-networks, investigated in \cite{ments,tents}, we extend ANTS to work in the more practical setting of online learning, shown in Algorithm~\ref{algo:learning}. The learning loop interleaves phases of data collection and training. When collecting an episode, we initialize the temperature to a constant value $\tau_0$, then pass and update it in the planner calls. We train the $Q$-network using SGD on samples from the replay buffer, with a loss function $L$ described in Section~\ref{sec:learning}.\textbf{}

\begin{algorithm}[ht]
\caption{Online learning.}
\label{algo:learning}

\begin{algorithmic}
\Function{learning\_loop}{}
\State $\textrm{replay\_buffer} \gets ()$
\While{\textbf{not} convergence}
\State episode $\gets \textsc{collect\_episode}$
\State $\textrm{replay\_buffer} \gets \textrm{replay\_buffer} \oplus (\textrm{episode})$
\State $\textrm{batch} \gets \textsc{replay\_sample}(\textrm{replay\_buffer})$
\State $\hat{Q} \gets \textsc{sgd}(L, \textrm{batch}, \hat{Q})$
\EndWhile
\EndFunction
\end{algorithmic}

\begin{algorithmic}
\Function{collect\_episode}{}
\State $s \gets s_0; \tau \gets \tau_0$
\State episode $\gets (s_0)$
\While{\textbf{not} $\textsc{is\_terminal}(s)$}
\State $a, \tau, q \gets \textsc{ants}(s, \tau)$
\State \textbf{sample} $s \sim \P(\cdot | s, a)$
\State episode $\gets$ episode $\oplus~(a, \tau, q, s)$
\EndWhile
\State \textbf{return} episode
\EndFunction
\end{algorithmic}

\end{algorithm}

The algorithms and equations in this paper refer to ANTS in combination with Shannon entropy \eqref{eq:shannon}. It is possible to extend MaxEnt MCTS to other types of entropy, as shown in \cite{tents}. We evaluate our method also in a setting with Tsallis entropy \eqref{eq:tsallis}; we denote the versions with Shannon and Tsallis entropy as ANTS-S and ANTS-T, respectively.


\subsection{Temperature adaptation} \label{sec:temp_adaptation}

In our method, the temperature $\tau$ is chosen considering entropies (depending on $\tau$) of nodes expanded by the MCTS planner (denoted by $\S_\mathrm{tree}$).  
Specifically, $\tau$ is varied so that the mean node entropy matches the prescribed value $\Havg$. This is achieved by finding a root of the following function:
\begin{align*}
    \H^+(\tau) &= \left[ \frac{1}{|\S_\mathrm{tree}|} \sum_{s \in \S_\mathrm{tree}} \H(\pi_\tau(\cdot | s)) \right] - \Havg, \\
    \textrm{where}~&\pi_\tau(a | s) \propto \exp \left( \frac{1}{\tau} Q(s, a) \right).
\end{align*}
%
%
Any root-finding algorithm can be applied here. For simplicity, we use the derivative-free Brent algorithm \cite{brent}. To avoid numerical issues when $\tau \approx 0$, we impose a lower bound on the temperature: $\tau_\mathrm{min}$. To avoid instabilities, we smooth the temperature using an exponential moving average. We smooth in $\log$-space to be able to quickly adapt the order of the magnitude. This results in the following two lines in Algorithm~\ref{algo:ants}:
\begin{align*}
    & \tau \gets \max(\textsc{find\_root}(\mathcal{H}^+), \tau_\mathrm{min}) \\
    & \Tilde{\tau} \gets \exp(\alpha \log \Tilde{\tau} + (1 - \alpha) \log \tau),
\end{align*}
where $\alpha$ is the parameter of the moving average.

\subsection{Pseudoreward shaping} \label{sec:shaping}
The $\textsc{backup}$ used in Algorithms~\ref{algo:ants} and~\ref{algo:select} is a modification of \eqref{eq:sqi}:
\begin{align} \label{eq:shaping}
    \begin{split}
        Q(s, a; \tau) & \leftarrow \R(s, a) + \gamma \Esp V(s'; \tau) \\
        V(s; \tau) & \leftarrow \tau \log \sum_{a \in \A} \exp \left( \frac{1}{\tau} Q(s, a; \tau) \right) - \tau\H_\mathrm{max},
    \end{split}
\end{align}

where $\H_\mathrm{max}$ is the maximum possible entropy. This modification is designed to prevent the inefficient depth-first behavior, which we observed in our experiments. Since values calculated by \eqref{eq:sqi} include positive entropy pseudorewards, they overestimate the true returns. As a result, the planner can start deepening the same, single path, because it keeps receiving high values along it. The pseudoreward shaping term $- \tau\H_\mathrm{max}$ in \eqref{eq:shaping} mitigates this effect. We note that as the subtracted value does not depend on the action, this shaping does not change the optimal policy obtained at infinite computational budget. 
\kucil{a) is there a more theoretical reason for that b) this is lack of shift invariance (we see something like this in vanilla PG, which is typically mitigated by introducing baseline)} \pk{proof is the same as for vanilla RL}

\subsection{$Q$-value estimation} \label{sec:value_estimation}

In the simulation phase, we estimate the $Q$-values of the newly-expanded leaves in the tree using a given $Q$-network. We initialize their $Q$-values as $Q(s, a) \leftarrow \hat{Q}(s, a; \tau)$, where $\hat{Q}$ denotes the Q-network. This initialization method is a great deal simpler than the one used in the original MaxEnt MCTS formulation \eqref{eq:tents_eval}. In addition to its simplicity, our method yields considerable improvement, as demonstrated in Section~\ref{sec:ablations}.
Another important element is accommodating temperature changes during planning. For this purpose, we inject information about the current temperature $\tau$ to the network - the $Q$-values are influenced by it. We have discovered that giving the network access to this information greatly stabilizes the algorithm.
\subsection{Online learning} \label{sec:learning}

\begin{figure*}[!ht]
 \centering
 \includegraphics[width=\textwidth]{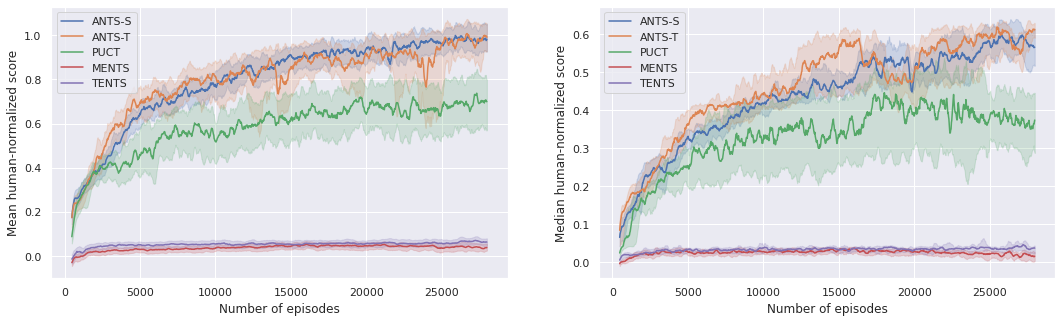}
\caption{Learning curves of the online learning experiments. Each plot represents the mean or median over the 21 games of the human-normalized score. Bold lines denote a median of five seeds. Shaded regions denote the interquartile range. Each point corresponds to a mean of 24 episodes. Each episode is run for a maximum of 1000 interactions. The curves are smoothed by a rolling mean of 10 points. \kucil{See "statistical precupice". Can you use to to produce intervals or distribution? (possibly a lot of work, but they cover exactly this setup) / they also propose other measure (iqm)}}
\label{fig:e2e}
\end{figure*}

The main evaluation setting considered in this paper is that of online learning. Our learning loop is similar to the one used in \cite{alphazero} and shown in Algorithm~\ref{algo:learning}. We emphasize that our algorithmic developments, presented above, and the learning details below were critical for making online learning feasible.

The loop consists of two phases: data collection and training. In the first phase, the planner is run for an episode, using a neural network for initializing the $Q$-values of leaves added to the tree. The episodes, along with state $Q$-values calculated by the planner, are stored in a replay buffer and used to train the neural network in the next phase. We note that the planner $Q$-values are typically more accurate than the ones from the network; this difference provides the learning signal.

During data collection, we use an exploration scheme similar to \eqref{eq:puct_sel}. Actions performed at each step in the environment are sampled from
\begin{equation} \label{eq:ants_sel}
    \pi(a) = \pi_{\tau \cdot \tau_\textrm{sel}}^\mathrm{e3w}(a | s_{\textrm{root}}),
\end{equation}
where $\tau_\textrm{sel} \in (0, 1)$ is an action selection temperature parameter. We use it to interpolate between the E3W \eqref{eq:e3w} and the argmax policy: $a(s) = \argmax_{a \in \A}~Q(s, a)$. The network is trained with L2 loss on the $Q$-value of the action performed in the collected experience: $L(\hat{Q}, s, a, \tau, q) = (\hat{Q}(s, a; \tau) - q)^2$

\iffalse

\subsection{Theoretical analysis}

The exponential convergence result for E2W in the softmax bandit problem, formulated and proven by \cite{ments}, still holds when the temperature changes, under the following conditions:
\begin{enumerate}
    \item The temperature $\tau_t$ in adaptation step $t$ converges to $\tau$ at rate $|\tau_t - \tau| = O(1 / \log t)$. We note that since it is a very slow convergence rate, this assumption is easily satisfied in practice.
    \item The reward distribution is sub-Gaussian. The same assumption is required for exponential convergence of E2W with constant temperature \cite{ments}.
\end{enumerate}
The formal statement of this result and its proof are described in \appendixref{app:theory}.

\fi

\fi

\section{Experiments}

In our main series of experiments, we plan using $Q$-networks trained online, as described in Section~\ref{sec:learning}. Then we present experiments with $Q$-networks pretrained using DQN, as done in \cite{ments} and \cite{tents}. Next, we provide an empirical comparison between two parameterizations of maximum entropy planning: via temperature and via entropy. Finally, we present ablations of various components of ANTS using pretrained $Q$-networks.

We present the results of ANTS combined with two types of entropy: ANTS-S using Shannon entropy \eqref{eq:shannon}, and ANTS-T using Tsallis entropy \eqref{eq:tsallis}. As baselines we include PUCT, MENTS, and TENTS. PUCT denotes MCTS with the PUCT selection strategy. We run our experiments on a set of $21$ Atari games and using the perfect model of the environment, in a setting similar to \cite{tents}. To ensure a fair comparison, we ran an extensive hyperparameter tuning of all presented methods, using equal computational budgets. The code and hyperparameter settings are available at \url{https://github.com/adaptive-entropy-tree-search/ants}.

\subsection{Main experiment: online learning} \label{sec:loop}

\kucil{If I understand correctly Fig 1 is the main figure, so put this on the same page as the section Experiments. Also say that is the main figure somewhere.} \pk{can't put it earlier than the first reference to the figure}

\kucil{reverse the order of sentences: first maxent mcts that works; outperforms / on par with puct-mcts; significantly outperforms ments and tents}

In this section we present experiments showing that ANTS is the first maximum entropy MCTS method competitive in the online learning setting. Specifically, ANTS significantly exceeds the results of the prior methods MENTS and TENTS. Moreover, it outperforms the state-of-the-art PUCT (utilized in AlphaZero and MuZero), at the same time being more stable.

As we can see in Figure~\ref{fig:e2e}, the performances of both ANTS-S and ANTS-T consistently improve as they collect more data. Both algorithms exhibit similar performance, though ANTS-T appears slightly more sample efficient, while ANTS-S learns more steadily and has lower variance between seeds. PUCT learns slowly and unsteadily, and its median performance eventually starts to deteriorate. MENTS and TENTS quickly plateau at a low score, which confirms the importance of the improvements made in ANTS in the online learning setting.

In these experiments, to collect diverse data for network training, we employ the exploration strategy \eqref{eq:ants_sel} for ANTS, MENTS, and TENTS, and \eqref{eq:puct_sel} for PUCT. The hyperparameter $\tau_\textrm{sel}$ is tuned for each method separately. We run five experiments for each algorithm and game, each collecting $28000$ episodes in total. An important technical step towards stabilizing training was providing the current temperature as an additional input to the Q-network, as described in Section~\ref{sec:value_estimation}.
 
\subsection{Pretrained Q-networks} \label{sec:pretrained}

\begin{table}[!ht]
\caption{Results on Atari using pretrained Q-networks}
\label{tab:pretrained}
\setlength{\tabcolsep}{3pt}
\begin{center}
\begin{small}
\begin{tabular}{lrrrrr}
\toprule
 & PUCT & MENTS & TENTS & ANTS-S & ANTS-T \\
\midrule
Alien    & \textbf{2853} & \textbf{2590} & \textbf{2606} & \textbf{2783} & 2554 \\
Amidar    & 137 & \textbf{251} & \textbf{236} & 148 & 184 \\
Asterix    & \textbf{48506} & 8305 & 10406 & \textbf{51741} & \textbf{49758} \\
Asteroids    & \textbf{3657} & \textbf{3786} & 3541 & \textbf{4059} & \textbf{3554} \\
Atlantis    & \textbf{278156} & 275681 & 267429 & \textbf{277870} & \textbf{280492} \\
BankHeist    & 680 & 543 & 472 & \textbf{707} & \textbf{715} \\
BeamRider    & \textbf{21745} & 9424 & 7534 & \textbf{21268} & 18536 \\
Breakout    & \textbf{389} & 306 & 243 & 351 & \textbf{377} \\
Centipede    & 18662 & 59167 & 99631 & 51387 & \textbf{133842} \\
DemonAttack    & \textbf{55947} & 49952 & 45960 & \textbf{53998} & \textbf{53462} \\
Enduro    & \textbf{794} & \textbf{800} & \textbf{800} & \textbf{794} & \textbf{800} \\
Frostbite    & \textbf{204} & 197 & 197 & 179 & 184 \\
Gopher    & 9606 & 9330 & 8150 & \textbf{10770} & \textbf{10766} \\
Hero    & \textbf{20715} & 19940 & 19845 & 20618 & 20443 \\
MsPacman    & 3880 & 3873 & 3358 & \textbf{4923} & \textbf{4846} \\
Phoenix    & 8875 & 6788 & 6857 & \textbf{9187} & \textbf{9557} \\
Qbert    & 15120 & 13877 & 13815 & \textbf{15472} & 15223 \\
Robotank    & \textbf{55} & 27 & 26 & 49 & 51 \\
Seaquest    & \textbf{3270} & 1764 & 1533 & 2557 & 2656 \\
SpaceInvaders    & \textbf{4630} & 2043 & 2024 & \textbf{4714} & \textbf{4453} \\
WizardOfWor    & \textbf{11834} & 7063 & 8030 & \textbf{12937} & \textbf{13386} \\
\midrule
\#best scores & \textbf{14/21} & 4/21 & 3/21 & \textbf{14/21} & 13/21 \\

\bottomrule
\end{tabular}
\end{small}
\end{center}
\end{table}

In this section we present experiments using pretrained Q-networks. This evaluation setting allows us to test the properties of the planners in isolation from learning. The results are shown in Table~\ref{tab:pretrained}. As we can see, ANTS-S performs strongly, achieving the best result or a tie on 14 out of 21 games, on others being close to the best result. Our highly tuned PUCT presents a challenging baseline, outperforming both MENTS and TENTS in almost every case. Both ANTS-S and ANTS-T perform significantly better than MENTS and TENTS, which provides evidence that the improvements made in ANTS are beneficial also when the Q-network is supplied externally. Interestingly, the performance of both variants of ANTS using pretrained networks is on-par with PUCT, whereas in the online learning experiments ANTS performed better. These results give rise to a hypothesis that the advantage of ANTS over PUCT in online learning is caused by providing better data for training neural networks, rather than better planning using a given neural network.

In Table~\ref{tab:pretrained} we report the mean of $100$ episodes. Each episode is run for $10^4$ interactions. For each environment, we bold the methods for which the difference with the highest score is not statistically significant (determined using a $t$-test with $p < 0.05$).

For those experiments, in ANTS, MENTS and TENTS we select an action with the maximum $Q$-value returned by the planner. In PUCT we select an action with the highest visitation count. Similar to \cite{ments,tents}, we use networks pretrained with DQN for leaf initialization. For this purpose, we utilize the publicly available checkpoints from the Dopamine package \cite{dopamine}.

\subsection{Robustness} \label{sec:robustness}

\begin{figure}[ht]
 \centering
 \includegraphics[width=0.47\textwidth]{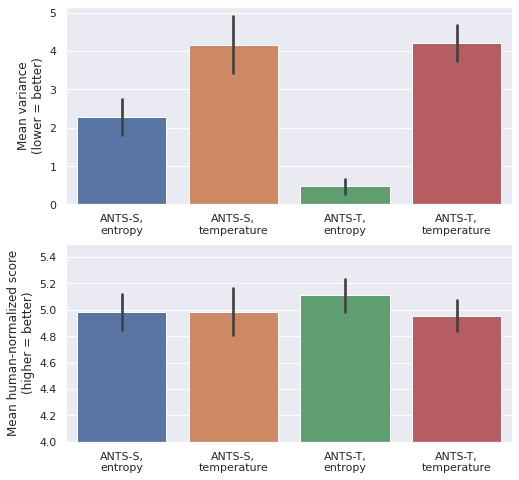}
\caption{Robustness experiments on the 21 Atari games. We average over 24 episodes for each variant and game. Error bars denote 95\% confidence intervals. \textit{ANTS-*, entropy} denote the full ANTS algorithm. \textit{ANTS-*, temperature} denote ANTS without temperature adaptation, keeping the temperature constant. These variants are equivalent to MENTS and TENTS with our simulation and backpropagation improvements described in Section~\ref{sec:method}.}
\label{fig:robustness}
\end{figure}

\piotrm{Proposition}

We now present a comparison of two parameterizations of maximum entropy planners: by mean entropy, used by ANTS, and by temperature, used by MENTS and TENTS. We evaluate the robustness to different parameter settings of the two parameterizations, as well as their performance when the hyperparameters are tuned. We measure the robustness by the \textit{mean variance} score, described below, with lower scores indicating higher robustness. The results are shown in Figure~\ref{fig:robustness}.

As we can see, ANTS parameterized by entropy is more robust than the variant parameterized by temperature. In the case of ANTS-S, mean variance of the entropy parameterization is 2 times lower than of the temperature parameterization. In the case of ANTS-T, the difference is much greater: over 8 times. In terms of the absolute performance, ANTS-T parameterized by entropy is noticeably better than its temperature-parameterized counterpart. In the case of ANTS-S, the difference is negligible.

Intuitively, an algorithm can be said to be robust to different hyperparameter choices, if its performance does not vary too much with different hyperparameter settings. To capture this intuition, we define our robustness metric as the variance of human-normalized scores attained at different hyperparameter values, where hyperparameters are chosen from a predefined set. We compute this metric separately for each game, then average it over games:
\kucil{maybe $sc\to \mathtt{score}$?} \pk{agreed}
\begin{equation}
    \rho(A, H) = \mathbb{E}_{g \sim G} \mathrm{Var}_{h \sim H}~\mathtt{score}(A, h, g),
\end{equation}
where $A$ indicates the algorithm using a given parameterization, $H$ is a set of its possible hyperparameter values, $G$ is the set of games, $\sim$ denotes sampling from the uniform distribution over a finite set, and $\mathtt{score}(A, h, g)$ is the score of algorithm $A$ using hyperparameter $h$ on game $g$.

\kucil{If we need more space, consider moving the details to the appendix} \pk{no appendix}

To ensure a fair comparison, the hyperparameter sets $H$ should be of equal size and span the range of sensible values. For the temperature parameterization, the range of possible temperature values is $(0, \infty)$. To handle this infinite interval we choose temperatures equidistant in log-space, with endpoints selected to be the minimum and maximum temperature that worked best on some game during hyperparameter tuning. In the case of entropy parameterization, the interval is $(0, \mathcal{H}_\textrm{max})$, where $\mathcal{H}_\textrm{max}$ depends on the entropy type and number of actions. We take values spanning this interval, equidistant in linear space.

\begin{figure}[ht]
 \centering
 \includegraphics[width=0.47\textwidth]{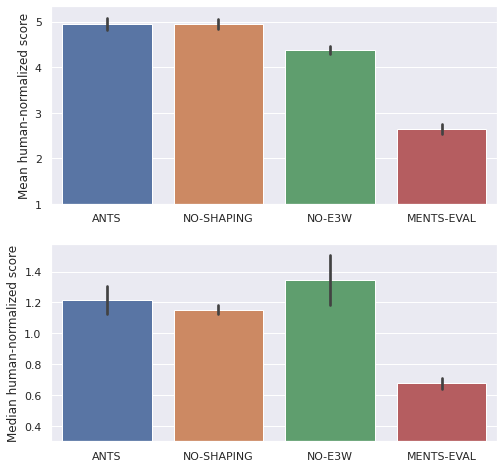}
\caption{Ablations of ANTS-S on the 21 Atari games. We average over 24 episodes for each variant and game. Error bars denote 95\% confidence intervals.}
\label{fig:ablations}
\end{figure}

\subsection{Ablations} \label{sec:ablations}

In order to measure the contribution of the different components of ANTS, we perform additional experiments with pretrained networks, removing the individual features of our method. For each variant we report the mean and median human-normalized score. The results are shown in Figure~\ref{fig:ablations}. ANTS \piotrm{We should perhaps write ANTS-S consistently; also on the figure} denotes the full algorithm using Shannon entropy. NO-SHAPING denotes ANTS without pseudoreward shaping and MENTS-EVAL denotes a variant of ANTS where leaf initialization is performed using \eqref{eq:tents_eval}, as done in \cite{ments} and \cite{tents}. We additionally include NO-E3W, denoting ANTS that samples directly from $\pi(a | s) \propto \exp \left( Q(s, a) / \tau \right)$ during the selection phase, instead of using E3W.

In Figure~\ref{fig:ablations} we can see that $Q$-value estimation using raw outputs of the Q-network plays the largest role out of the evaluated components of ANTS. One possible explanation is that leaf initialization using formula \eqref{eq:tents_eval} retains only information about relations between actions, losing information about the absolute value of a state. This could prevent the planner from a meaningful comparison of nodes higher in the tree. In the case of pseudoreward shaping, the difference is less apparent and only visible in the median human-normalized score plot. Interestingly, NO-E3W performs worse than ANTS in terms of mean score and better in terms of median score. It appears that E3W sampling is necessary on a small number of games, which escalates the mean score, but hurts on most games, which diminishes the median score. It is also worth noticing that the median NO-E3W score has considerably higher variance.

The ablations use the same base hyperparameter as the experiments in Section~\ref{sec:pretrained}. In the MENTS-EVAL experiment, we re-tune $\tau_\textrm{init}$ jointly on all games, maximizing the mean human-normalized score. In both cases, we use the same set of hyperparameter values as when tuning MENTS.

\section{Related Work}

\kucil{Shorten ideally to 1/2 of a column. A lot of the references already covered earlier. This was one of Tony's tips for us, that there is this trend that people think related work takes too much space and only the most relevant (directly related) works should be cited. Otherwise, didn't read this section}

\pk{move; discuss the differences}

\textbf{Reinforcement Learning} Recent years have brought significant advancements in reinforcement learning due to the use of powerful deep neural network (DNN) function approximators. \cite{dqn} introduce DQN: Q-learning with DNNs, leading to a human-level performance on the Arcade Learning Environment \cite{ale}. The algorithmic progress has been perhaps equally important. \cite{rainbow} combine multiple improvements to DQN, enabling superior performance across the ALE benchmark. \cite{agent57} advance even further, leading to an algorithm that achieves superhuman performance on all 57 ALE tasks.
%

\textbf{Monte Carlo Tree Search} MCTS is a combination of the UCB strategy for multi-armed bandits with tree search  \cite{uct}. \cite{mcts-survey} provide a survey of methods in the MCTS family. \cite{alphazero} introduce AlphaZero - MCTS powered with neural networks and self-play, to obtain a system achieving superhuman performance in complex strategic board games: chess, shogi and go. \cite{muzero} develop MuZero, an extension of AlphaZero with a learned model, and establish a new state-of-the-art on ALE.\cite{mcts-as-rpo} discover a link between MCTS and policy optimization and derive a novel selection strategy.



\textbf{Maximum Entropy RL} The Principle of Maximum Entropy (PME) has shown success in modeling human behavior \cite{maxent-phd}. \cite{control-as-inference} describe the maximum entropy RL problem as inference in a probabilistic graphical model. \cite{maxent-is-the-answer} analyze the maximum entropy RL problem and show its equivalence to classes of MDPs with uncertain or adversarial reward functions. \cite{pcl} establish a connection between policy-based and value-based RL under entropy regularization. \cite{sql} introduce Soft Q-Learning: a deep RL method derived from soft Q-iteration \eqref{eq:sqi}. \cite{sac} present another maximum entropy RL method: Soft Actor-Critic (SAC), achieving state-of-the-art performance in continuous control tasks. \cite{ments} use the maximum entropy framework to derive E2W: an optimal sampling strategy for the softmax stochastic bandit problem and introduce MENTS - an MCTS algorithm using this strategy in the selection phase. \cite{tents} extend E2W to different types of entropy: Tsallis entropy and relative Shannon entropy, resulting in the E3W sampling strategy. Our work differs from both \cite{ments} and \cite{tents} by the introduction of temperature adaptation, simple $Q$-value estimation, and pseudoreward shaping.


\section{Conclusions}

\pk{add learning a model to future work? or maybe we shouldn't draw too much attention to this?}\piotrm{Honest and comprehensive 'limitation' section usually makes a good impression. I've seen a few examples of reviewers explictly prising such.} \pk{no space, unfortunately}

\kucil{I agree with PM: reviewers always were pleased about this (and neurips template suggests this). I suggest multiple cuts earlier, so there should be a room for that}

\kucil{If you agree with comments concernign the storyline, the text below should be revised}

In this work, we have presented Adaptive Entropy Tree Search: the first maximum entropy MCTS method competitive in the online learning setting. It delivers superior performance compared to the prior MaxEnt MCTS algorithms MENTS and TENTS, and outperforms the state-of-the-art PUCT. Moreover, ANTS is on par with PUCT when used as a standalone planner, with pretrained neural networks. Importantly, ANTS considerably improves upon the related MENTS and TENTS algorithms in both aspects. We have also demonstrated that the entropy reparameterization introduced by ANTS leads to increased robustness to hyperparameter settings. This makes ANTS particularly viable in real-world applications, where hyperparameter tuning is expensive. Finally, we have verified the design choices in a comprehensive ablation study.

We believe that a promising direction for future research is further analysis of maximum entropy approaches in combination with online learning. The high performance of ANTS in this setting can be attributed to good exploration or exploitation. Both can play a role - on the one hand, maximum entropy planning can be seen as an alternative method of exploration to UCT/PUCT; on the other hand, maximum entropy RL approaches are believed to aid in smoothing the optimization landscape of neural networks \cite{entropy-landscape}. We believe that this line of research would serve as a step towards developing practical planning approaches to challenging real-world problems.

\section*{Acknowledgments}

We thank Łukasz Kuciński and Konrad Czechowski for their helpful suggestions on the writing. The work of Piotr Kozakowski was supported by the Polish National Science Center grant 2020/37/N/ST6/04113. The work of Mikołaj Pacek was supported by the Digital Poland Operational Programme, financed by the European Structural and Investment Fund, as part of the project AI Tech (POPC.03.02.00-IP.01-00-006/20). The work of Piotr Miłoś was supported by the Polish National Science Center grant 2017/26/E/ST6/00622. This research was supported by the PL-Grid Infrastructure. Our experiments were managed using \url{https://neptune.ai}. We thank the Neptune team for providing us access to the team version and technical support.

\kucil{References formatting is buggy at some places}

\bibliographystyle{IEEEtran}
\bibliography{IEEEabrv,ants}

\begin{thebibliography}{10}
\providecommand{\url}[1]{#1}
\csname url@samestyle\endcsname
\providecommand{\newblock}{\relax}
\providecommand{\bibinfo}[2]{#2}
\providecommand{\BIBentrySTDinterwordspacing}{\spaceskip=0pt\relax}
\providecommand{\BIBentryALTinterwordstretchfactor}{4}
\providecommand{\BIBentryALTinterwordspacing}{\spaceskip=\fontdimen2\font plus
\BIBentryALTinterwordstretchfactor\fontdimen3\font minus
  \fontdimen4\font\relax}
\providecommand{\BIBforeignlanguage}[2]{{%
\expandafter\ifx\csname l@#1\endcsname\relax
\typeout{** WARNING: IEEEtran.bst: No hyphenation pattern has been}%
\typeout{** loaded for the language `#1'. Using the pattern for}%
\typeout{** the default language instead.}%
\else
\language=\csname l@#1\endcsname
\fi
#2}}
\providecommand{\BIBdecl}{\relax}
\BIBdecl

\bibitem{hassabis2017neuroscience}
D.~Hassabis, D.~Kumaran, C.~Summerfield, and M.~Botvinick,
  ``Neuroscience-inspired artificial intelligence,'' \emph{Neuron}, vol.~95,
  no.~2, pp. 245--258, 2017.

\bibitem{russell2002artificial}
S.~Russell and P.~Norvig, \emph{Artificial Intelligence: A Modern Approach},
  3rd~ed.\hskip 1em plus 0.5em minus 0.4em\relax USA: Prentice Hall Press,
  2009.

\bibitem{DBLP:journals/eor/BengioLP21}
\BIBentryALTinterwordspacing
Y.~Bengio, A.~Lodi, and A.~Prouvost, ``Machine learning for combinatorial
  optimization: A methodological tour d’horizon,'' \emph{European Journal of
  Operational Research}, vol. 290, no.~2, pp. 405--421, 2021. [Online].
  Available:
  \url{https://www.sciencedirect.com/science/article/pii/S0377221720306895}
\BIBentrySTDinterwordspacing

\bibitem{alphazero}
D.~Silver, T.~Hubert, J.~Schrittwieser, I.~Antonoglou, T.~Graepel,
  T.~Lillicrap, K.~Simonyan, and D.~Hassabis, ``{A general reinforcement
  learning algorithm that masters chess, shogi, and Go through self-play},''
  \emph{Science}, vol. 1144, pp. 1140--1144, 2018.

\bibitem{Agostinelli:2019aa}
\BIBentryALTinterwordspacing
F.~Agostinelli, S.~McAleer, A.~Shmakov, and P.~Baldi, ``Solving the rubik's
  cube with deep reinforcement learning and search,'' \emph{Nature Machine
  Intelligence}, vol.~1, no.~8, pp. 356--363, 2019. [Online]. Available:
  \url{https://doi.org/10.1038/s42256-019-0070-z}
\BIBentrySTDinterwordspacing

\bibitem{mcts-survey}
C.~Browne, E.~Powley, D.~Whitehouse, S.~Lucas, P.~Cowling, P.~Rohlfshagen,
  S.~Tavener, D.~Perez~Liebana, S.~Samothrakis, and S.~Colton, ``A survey of
  monte carlo tree search methods,'' \emph{IEEE Transactions on Computational
  Intelligence and AI in Games}, vol. 4:1, pp. 1--43, 03 2012.

\bibitem{uct}
L.~Kocsis and C.~Szepesvári, ``Bandit based monte-carlo planning,'' in
  \emph{In: ECML-06. Number 4212 in LNCS}.\hskip 1em plus 0.5em minus
  0.4em\relax Springer, 2006, pp. 282--293.

\bibitem{mogo}
C.-S. Lee, M.-H. Wang, G.~Chaslot, J.-B. Hoock, A.~Rimmel, O.~Teytaud, S.-R.
  Tsai, S.-C. Hsu, and T.-P. Hong, ``The computational intelligence of mogo
  revealed in taiwan's computer go tournaments,'' \emph{Computational
  Intelligence and AI in Games, IEEE Transactions on}, vol.~1, pp. 73 -- 89, 04
  2009.

\bibitem{mcts-security}
Y.~{Tanabe}, K.~{Yoshizoe}, and H.~{Imai}, ``A study on security evaluation
  methodology for image-based biometrics authentication systems,'' in
  \emph{2009 IEEE 3rd International Conference on Biometrics: Theory,
  Applications, and Systems}, 2009, pp. 1--6.

\bibitem{mcts-physics}
C.~Mansley, A.~Weinstein, and M.~L. Littman, ``Sample-based planning for
  continuous action markov decision processes,'' in \emph{Proceedings of the
  Twenty-First International Conference on International Conference on
  Automated Planning and Scheduling}, ser. ICAPS'11.\hskip 1em plus 0.5em minus
  0.4em\relax AAAI Press, 2011, p. 335–338.

\bibitem{mcts-production}
G.~Chaslot, S.~Jong, J.-T. Saito, and J.~Uiterwijk, ``Monte-carlo tree search
  in production management problems,'' \emph{Belgian/Netherlands Artificial
  Intelligence Conference}, 01 2006.

\bibitem{mcts-creative}
T.~Mahlmann, J.~Togelius, and G.~N. Yannakakis, ``Towards procedural strategy
  game generation: Evolving complementary unit types,'' in \emph{Applications
  of Evolutionary Computation}, C.~Di~Chio, S.~Cagnoni, C.~Cotta, M.~Ebner,
  A.~Ek{\'a}rt, A.~I. Esparcia-Alc{\'a}zar, J.~J. Merelo, F.~Neri, M.~Preuss,
  H.~Richter, J.~Togelius, and G.~N. Yannakakis, Eds.\hskip 1em plus 0.5em
  minus 0.4em\relax Berlin, Heidelberg: Springer Berlin Heidelberg, 2011, pp.
  93--102.

\bibitem{muzero}
J.~Schrittwieser, I.~Antonoglou, T.~Hubert, K.~Simonyan, L.~Sifre, S.~Schmitt,
  A.~Guez, E.~Lockhart, D.~Hassabis, T.~Graepel, T.~Lillicrap, and D.~Silver,
  ``Mastering atari, go, chess and shogi by planning with a learned model,''
  \emph{Nature}, vol. 588, pp. 604--609, 12 2020.

\bibitem{mcts-as-rpo}
\BIBentryALTinterwordspacing
J.-B. Grill, F.~Altch{\'e}, Y.~Tang, T.~Hubert, M.~Valko, I.~Antonoglou, and
  R.~Munos, ``{M}onte-{C}arlo tree search as regularized policy optimization,''
  in \emph{Proceedings of the 37th International Conference on Machine
  Learning}, ser. Proceedings of Machine Learning Research, H.~D. III and
  A.~Singh, Eds., vol. 119.\hskip 1em plus 0.5em minus 0.4em\relax PMLR, 13--18
  Jul 2020, pp. 3769--3778. [Online]. Available:
  \url{http://proceedings.mlr.press/v119/grill20a.html}
\BIBentrySTDinterwordspacing

\bibitem{effzero}
W.~Ye, S.~Liu, T.~Kurutach, P.~Abbeel, and Y.~Gao, ``Mastering atari games with
  limited data,'' in \emph{NeurIPS}, 2021.

\bibitem{maxent-phd}
\BIBentryALTinterwordspacing
B.~D. Ziebart, ``Modeling purposeful adaptive behavior with the principle of
  maximum causal entropy,'' Jul 2018. [Online]. Available:
  \url{https://kilthub.cmu.edu/articles/thesis/Modeling\_Purposeful\_Adaptive\_Behavior\_with\_the\_Principle\_of\_Maximum\_Causal\_Entropy/6720692/1}
\BIBentrySTDinterwordspacing

\bibitem{maxent-collective}
\BIBentryALTinterwordspacing
A.~Hernando, R.~Hernando, A.~Plastino, and A.~R. Plastino, ``The workings of
  the maximum entropy principle in collective human behaviour,'' \emph{Journal
  of The Royal Society Interface}, vol.~10, no.~78, p. 20120758, Jan 2013.
  [Online]. Available: \url{http://dx.doi.org/10.1098/rsif.2012.0758}
\BIBentrySTDinterwordspacing

\bibitem{maxent-is-the-answer}
B.~Eysenbach and S.~Levine, ``If maxent rl is the answer, what is the
  question?'' 2019.

\bibitem{sql}
\BIBentryALTinterwordspacing
T.~Haarnoja, H.~Tang, P.~Abbeel, and S.~Levine, ``Reinforcement learning with
  deep energy-based policies,'' in \emph{Proceedings of the 34th International
  Conference on Machine Learning}, ser. Proceedings of Machine Learning
  Research, D.~Precup and Y.~W. Teh, Eds., vol.~70.\hskip 1em plus 0.5em minus
  0.4em\relax International Convention Centre, Sydney, Australia: PMLR, 06--11
  Aug 2017, pp. 1352--1361. [Online]. Available:
  \url{http://proceedings.mlr.press/v70/haarnoja17a.html}
\BIBentrySTDinterwordspacing

\bibitem{sac}
\BIBentryALTinterwordspacing
T.~Haarnoja, A.~Zhou, P.~Abbeel, and S.~Levine, ``Soft actor-critic: Off-policy
  maximum entropy deep reinforcement learning with a stochastic actor,'' in
  \emph{Proceedings of the 35th International Conference on Machine Learning},
  ser. Proceedings of Machine Learning Research, J.~Dy and A.~Krause, Eds.,
  vol.~80.\hskip 1em plus 0.5em minus 0.4em\relax Stockholmsmässan, Stockholm
  Sweden: PMLR, 10--15 Jul 2018, pp. 1861--1870. [Online]. Available:
  \url{http://proceedings.mlr.press/v80/haarnoja18b.html}
\BIBentrySTDinterwordspacing

\bibitem{ments}
\BIBentryALTinterwordspacing
C.~Xiao, R.~Huang, J.~Mei, D.~Schuurmans, and M.~M\"{u}ller, ``Maximum entropy
  monte-carlo planning,'' in \emph{Advances in Neural Information Processing
  Systems}, vol.~32.\hskip 1em plus 0.5em minus 0.4em\relax Curran Associates,
  Inc., 2019, pp. 9520--9528. [Online]. Available:
  \url{https://proceedings.neurips.cc/paper/2019/file/7ffb4e0ece07869880d51662a2234143-Paper.pdf}
\BIBentrySTDinterwordspacing

\bibitem{tents}
T.~Dam, C.~D'Eramo, J.~Peters, and J.~Pajarinen, ``Convex regularization in
  monte-carlo tree search,'' 2020.

\bibitem{ale}
\BIBentryALTinterwordspacing
M.~G. Bellemare, Y.~Naddaf, J.~Veness, and M.~Bowling, ``The arcade learning
  environment: An evaluation platform for general agents,'' \emph{Journal of
  Artificial Intelligence Research}, vol. Vol. 47, pp. 253--279, 2012, cite
  arxiv:1207.4708. [Online]. Available: \url{http://arxiv.org/abs/1207.4708}
\BIBentrySTDinterwordspacing

\bibitem{branching-factor}
D.~Jurafsky and J.~H. Martin,
  \url{https://web.stanford.edu/~jurafsky/slp3/3.pdf}, 2020, accessed:
  2021-09-09.

\bibitem{sparse-mdp}
\BIBentryALTinterwordspacing
K.~Lee, S.~Choi, and S.~Oh, ``Sparse markov decision processes with causal
  sparse tsallis entropy regularization for reinforcement learning,''
  \emph{{IEEE} Robotics Autom. Lett.}, vol.~3, no.~2, pp. 1466--1473, 2018.
  [Online]. Available: \url{https://doi.org/10.1109/LRA.2018.2800085}
\BIBentrySTDinterwordspacing

\bibitem{brent}
R.~Brent, \emph{Algorithms for minimization without derivatives}.\hskip 1em
  plus 0.5em minus 0.4em\relax Prentice-Hall, 1973.

\bibitem{dopamine}
P.~S. Castro, S.~Moitra, C.~Gelada, S.~Kumar, and M.~G. Bellemare, ``Dopamine:
  A research framework for deep reinforcement learning,'' 2018.

\bibitem{dqn}
\BIBentryALTinterwordspacing
V.~Mnih, K.~Kavukcuoglu, D.~Silver, A.~A. Rusu, J.~Veness, M.~G. Bellemare,
  A.~Graves, M.~Riedmiller, A.~K. Fidjeland, G.~Ostrovski, S.~Petersen,
  C.~Beattie, A.~Sadik, I.~Antonoglou, H.~King, D.~Kumaran, D.~Wierstra,
  S.~Legg, and D.~Hassabis, ``Human-level control through deep reinforcement
  learning,'' \emph{Nature}, vol. 518, no. 7540, pp. 529--533, Feb. 2015.
  [Online]. Available: \url{http://dx.doi.org/10.1038/nature14236}
\BIBentrySTDinterwordspacing

\bibitem{rainbow}
\BIBentryALTinterwordspacing
M.~Hessel, J.~Modayil, H.~van Hasselt, T.~Schaul, G.~Ostrovski, W.~Dabney,
  D.~Horgan, B.~Piot, M.~G. Azar, and D.~Silver, ``Rainbow: Combining
  improvements in deep reinforcement learning.'' in \emph{AAAI}, S.~A.
  McIlraith and K.~Q. Weinberger, Eds.\hskip 1em plus 0.5em minus 0.4em\relax
  AAAI Press, 2018, pp. 3215--3222. [Online]. Available:
  \url{http://dblp.uni-trier.de/db/conf/aaai/aaai2018.html\#HesselMHSODHPAS18}
\BIBentrySTDinterwordspacing

\bibitem{agent57}
\BIBentryALTinterwordspacing
A.~P. Badia, B.~Piot, S.~Kapturowski, P.~Sprechmann, A.~Vitvitskyi, Z.~D. Guo,
  and C.~Blundell, ``Agent57: Outperforming the {A}tari human benchmark,'' in
  \emph{Proceedings of the 37th International Conference on Machine Learning},
  ser. Proceedings of Machine Learning Research, H.~D. III and A.~Singh, Eds.,
  vol. 119.\hskip 1em plus 0.5em minus 0.4em\relax PMLR, 13--18 Jul 2020, pp.
  507--517. [Online]. Available:
  \url{http://proceedings.mlr.press/v119/badia20a.html}
\BIBentrySTDinterwordspacing

\bibitem{control-as-inference}
S.~Levine, ``Reinforcement learning and control as probabilistic inference:
  Tutorial and review,'' 2018.

\bibitem{pcl}
\BIBentryALTinterwordspacing
O.~Nachum, M.~Norouzi, K.~Xu, and D.~Schuurmans, ``Bridging the gap between
  value and policy based reinforcement learning,'' in \emph{Advances in Neural
  Information Processing Systems}, I.~Guyon, U.~V. Luxburg, S.~Bengio,
  H.~Wallach, R.~Fergus, S.~Vishwanathan, and R.~Garnett, Eds., vol.~30.\hskip
  1em plus 0.5em minus 0.4em\relax Curran Associates, Inc., 2017. [Online].
  Available:
  \url{https://proceedings.neurips.cc/paper/2017/file/facf9f743b083008a894eee7baa16469-Paper.pdf}
\BIBentrySTDinterwordspacing

\bibitem{entropy-landscape}
\BIBentryALTinterwordspacing
Z.~Ahmed, N.~Le~Roux, M.~Norouzi, and D.~Schuurmans, ``Understanding the impact
  of entropy on policy optimization,'' in \emph{Proceedings of the 36th
  International Conference on Machine Learning}, ser. Proceedings of Machine
  Learning Research, K.~Chaudhuri and R.~Salakhutdinov, Eds., vol.~97.\hskip
  1em plus 0.5em minus 0.4em\relax PMLR, 09--15 Jun 2019, pp. 151--160.
  [Online]. Available: \url{http://proceedings.mlr.press/v97/ahmed19a.html}
\BIBentrySTDinterwordspacing

\end{thebibliography}

\end{document}